\documentclass[letterpaper, 10 pt, conference]{ieeeconf}  

\IEEEoverridecommandlockouts                              

\usepackage[utf8]{inputenc} 
\usepackage[T1]{fontenc}    
\usepackage{hyperref}       
\usepackage{url}            
\usepackage{booktabs}       
\usepackage{amsfonts}       
\usepackage{nicefrac}       
\usepackage{microtype}      
\usepackage{xcolor}         
\usepackage{graphicx}
\usepackage{multirow}
\usepackage{subcaption}
\usepackage{amsmath} 
\usepackage{cuted}
\usepackage{capt-of}
\usepackage{marvosym}
\title{\LARGE \bf
Token-World: World Modeling in Vision-Language Model Token Space for Robot Manipulation
}

\author{
Chuyao Fu$^{1,2,6*}$,
Xiaowei Chi$^{3,6*\dagger}$,
Yuhan Rui$^{2,6}$,
Yu-Kai Wang$^{1,6}$,
Zezhong Qian$^{1}$,
Xiaojie Zhang$^{3}$, \\
Yunfan Lou$^{1}$,
Kevin Zhang$^{1,6}$,
Kuangzhi Ge$^{1,6}$,
Chak Wing Mak$^{1,6}$,
Zhiyang Chen$^{1,6}$, \\
Athena Zhuoming Zhong$^{5}$,
Hongyang Cheng$^{1}$,
Haoran Li$^{4}$,
Yike Guo$^{3}$,
Sirui Han$^{3}$,
Shanghang Zhang$^{1}$\textsuperscript{\Letter}
\thanks{\textsuperscript{*} Equal Contribution.
\textsuperscript{$\dagger$} Project Leader.
\Letter\ Corresponding author. Correspondence to:
Shanghang Zhang
\href{mailto:shanghang@pku.edu.cn}{<shanghang@pku.edu.cn>}.
This work was supported by the National Natural Science Foundation of China
(62476011) and the Beijing Natural Science Foundation (L252060).}
\\[0.35em]
\small
$^{1}$State Key Laboratory of Multimedia Information Processing,
School of Computer Science, Peking University
\\[-0.1em]
$^{2}$Southern University of Science and Technology
\\[-0.1em]
$^{3}$Hong Kong University of Science and Technology
\quad
$^{4}$Institute of Automation, Chinese Academy of Sciences
\\[-0.1em]
$^{5}$University of Pennsylvania
\quad
$^{6}$MUKA Robotics
}

\begin{document}

\maketitle

\vspace{-12mm}

\begin{strip}
    \centering
    \includegraphics[width=0.8\textwidth]{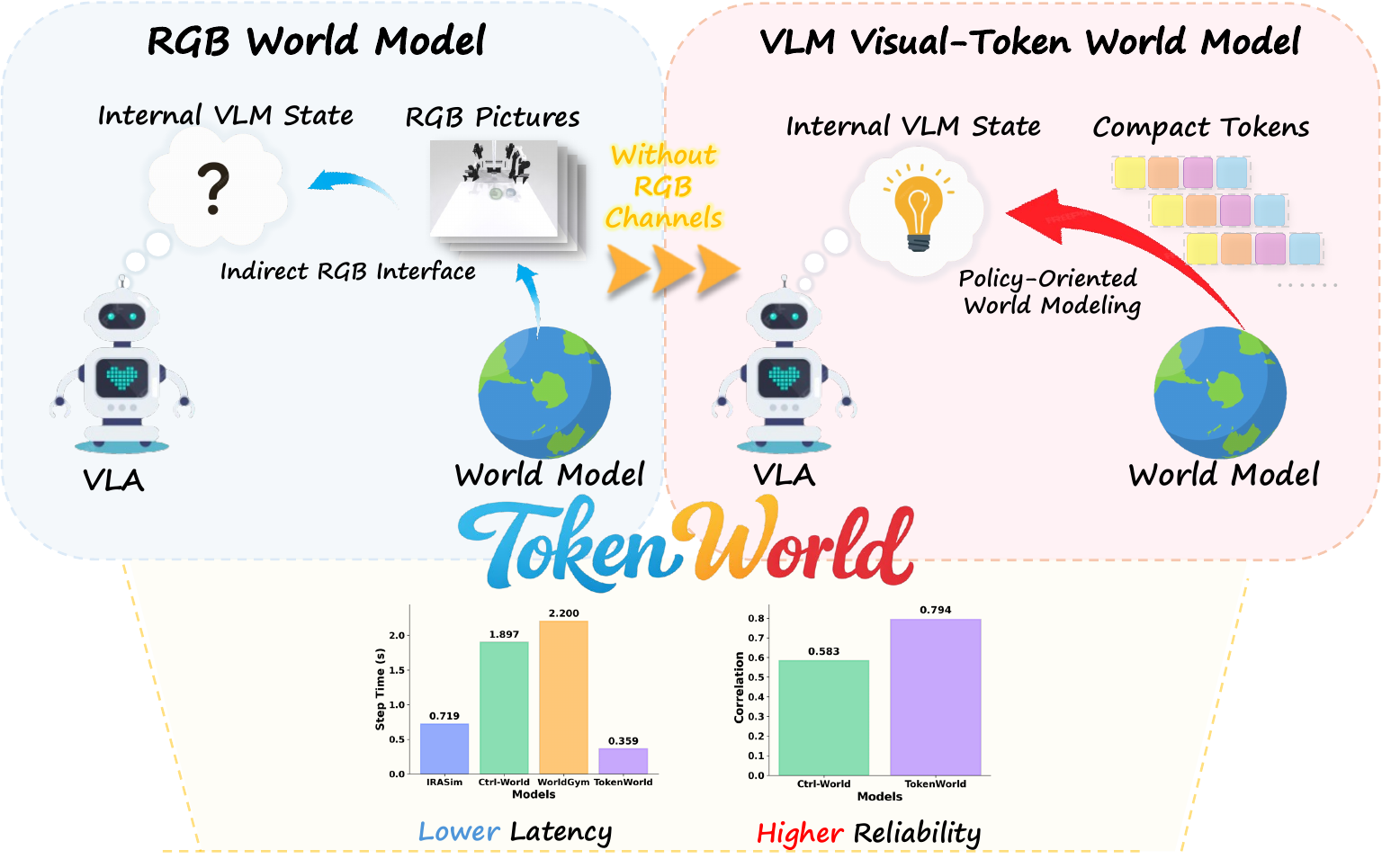}
    \vspace{-1.5mm}
    \captionof{figure}{
    \textbf{Token-World simulates directly in compact VLM visual-token space.}
    Unlike RGB-based world models that generate future images and re-encode them into policy inputs, Token-World models action-conditioned dynamics directly in compact VLM tokens.
    This design avoids intermediate RGB generation, reduces simulator step time to $0.359$\,s, and yields more reliable closed-loop policy evaluation ($r=0.794$ vs.\ $0.583$ for Ctrl-World).
    }
    \label{fig:teaser}
\end{strip}

\thispagestyle{empty}
\pagestyle{empty}

\begin{abstract}

A common approach to world-model simulation for vision-language-action (VLA) systems is to predict future RGB observations and then re-encode them into policy inputs, introducing an indirect interface between simulation and downstream policy execution. We instead investigate whether world dynamics can be modeled in a compact, policy-oriented state derived from VLM visual tokens. A key challenge is that raw VLM visual tokens are high-dimensional, making efficient and accurate autoregressive dynamics modeling challenging. To address this, we introduce Token-World, an action-conditioned world model that compresses VLM features into a compact token state, learns future dynamics in this reduced space, and maps predicted states back to the original policy-facing representation for downstream use. Across manipulation benchmarks, Token-World improves open-loop feature fidelity and policy-action consistency over recent world-model simulators, with slower degradation over long rollout horizons. In closed-loop evaluation, its simulated policy performance correlates more strongly with reference policy performance than Ctrl-World ($r=0.794$ vs.\ $0.583$), while requiring lower simulation latency. Ablations further show that compact-representation design and dimensionality substantially affect future-state prediction. Code will be available at \url{https://chuyaofu.github.io/Token-World/}.

\end{abstract}

\section{INTRODUCTION}

World models \cite{ha2018world} offer a promising route toward scalable embodied learning by serving as learned simulators of action-conditioned environment dynamics \cite{finn2017deep,hafner2019dream,hafner2019learning,wu2023daydreamer}. Instead of executing every behavior in the physical world, an agent can roll out future observations inside the model, which is especially valuable in robotics where real-world interaction is costly and slow\cite{wu2023daydreamer}. Recent work has therefore explored world models as simulators from several perspectives: using predicted futures to evaluate policies \cite{quevedo2025worldgym, wang2026echoarena}, synthesizing additional interaction data for policy learning\cite{guo2025ctrl,chi2025wow}, and performing reinforcement learning through imagined rollouts\cite{zhu2025wmpo,jiang2025world4rl,jiang2026wovr}.

As world models are increasingly used as learned simulators, they are often paired with VLA policies that map visual observations and language instructions to actions \cite{team2024octo,kim2024openvla,pi0,pi05}. However, existing pipelines often use an indirect simulation interface, where future RGB observations are first predicted by the world model, then re-encoded into VLA visual tokens before being consumed by the VLA policy \cite{quevedo2025worldgym,guo2025ctrl,zhu2025wmpo,jiang2025world4rl,jiang2026wovr}. As world models move toward scalable simulators for VLA agents, this indirect interface becomes a critical bottleneck: the simulator predicts human-viewable pixels, whereas the downstream policy ultimately consumes policy-facing visual tokens.This mismatch is not merely computational: RGB reconstruction encourages the simulator to allocate capacity to visual details that may be irrelevant to the policy, while token-space rollout directly targets the representation used for downstream action generation.

A representation-aligned alternative is to simulate the future directly in the VLM visual-token space consumed by the policy. Such an interface avoids repeatedly reconstructing RGB observations and re-encoding them into policy inputs during imagined rollouts. However, policy-facing VLM tokens are high-dimensional representations optimized primarily for perception and action rather than generative dynamics modeling~\cite{kim2024openvla,pi0,pi05}. Directly predicting their temporal evolution therefore imposes both a substantial modeling burden and considerable computational cost.

In this work, we introduce \textbf{Token-World}, an autoregressive action-conditioned world-model simulator that rolls out in compact policy-aligned VLM token space. Its dynamics model is a flow-matching DiT, without relying on pretrained video-generation models. To make high-dimensional policy-facing VLM features tractable for dynamics learning, Token-World compresses the feature dimension into a compact semantic token state. Future states are predicted in this compact space and mapped back to the original VLM visual-token space only when consumed by the downstream policy, avoiding intermediate RGB generation throughout the rollout.

Across manipulation benchmarks, Token-World improves open-loop feature fidelity and policy-action consistency over recent world-model simulators, while degrading more slowly over long rollout horizons. 
In closed-loop evaluation, its simulated success rates track reference policy performance more closely than Ctrl-World ($r=0.794$ vs.\ $0.583$), while providing higher simulation efficiency. 
Controlled ablations further show that both the design and dimensionality of the compact VLM representation substantially affect future-state prediction.
In summary, our contributions are threefold:
\begin{itemize}
    \item We study direct VLM visual-token simulation as an alternative to RGB-based world-model pipelines, enabling the simulator to operate on the representation interface consumed by downstream VLA policies.

    \item We present Token-World, an autoregressive action-conditioned world-model simulator that uses a flow-matching DiT to model dynamics in a compact VLM visual-token space, without relying on pretrained video-generation models.

    \item We validate Token-World through open-loop prediction, closed-loop policy evaluation, and representation ablations, showing improved long-horizon feature and policy-action fidelity, more reliable policy success estimation, lower simulation cost, and the importance of compact-representation design for dynamics prediction.
\end{itemize}

\section{Related Work}

\subsection{World Models as Learned Simulators}

World models learn action-conditioned dynamics to support planning and policy learning through imagined rollouts \cite{ha2018world,hafner2019dream,hafner2019learning,wu2023daydreamer}. Recent robotic world models have been used for policy evaluation \cite{li2025worldeval,quevedo2025worldgym}, synthetic interaction generation \cite{guo2025ctrl,jang2025dreamgen}, and policy optimization \cite{zhu2025wmpo}. Token-World focuses on a complementary question: what representation should serve as the state of a world model coupled with a VLA policy?

\subsection{Representation Design for Latent World Models}

Latent world models predict future states in learned representation spaces.
DINO-WM and related methods model pretrained visual features
\cite{zhou2024dino,karypidis2024dino,baldassarre2025back},
while JEPA-style models predict future latent embeddings
\cite{assran2025v,bardes2024revisiting}.
Beyond generic feature latents, Mask World Model predicts future semantic masks
as a structured geometric bottleneck that suppresses task-irrelevant appearance
variation \cite{lou2026mask}.
Recent VLA-oriented methods move prediction closer to policy representations:
DIAL predicts futures in VLM feature space \cite{chen2026dial},
LaWAM uses latent visual subgoals \cite{chen2026lawam},
and LaST$_0$ reasons through token-efficient latent spatio-temporal states \cite{liu2026last}.
Together, these works highlight the importance of the predictive representation;
our focus is instead on how a compact dynamics state should be constructed from
high-dimensional policy-facing representations for action-conditioned world modeling.

Related generative models have also studied representation design for diffusion.
Representation Autoencoders use pretrained semantic features as diffusion latents
\cite{zheng2026diffusion}, while PS-VAE maps high-dimensional representation features into a compact, KL-regularized latent space for image generation and editing
\cite{zhang2025both}. Compression has also appeared in world models:
DeltaWorld compresses DINO feature changes into a delta token
\cite{kerssies2026frame}, and OneWM-VLA compresses each visual view into a semantic token
\cite{tang2026one}. Token-World instead compresses policy-facing VLM tokens along the feature dimension while preserving their spatial layout, distinguishing it from approaches that aggregate frame- or view-level information into a small number of compact tokens.

\begin{figure*}[t]
\centering
\includegraphics[width=1.0\linewidth]{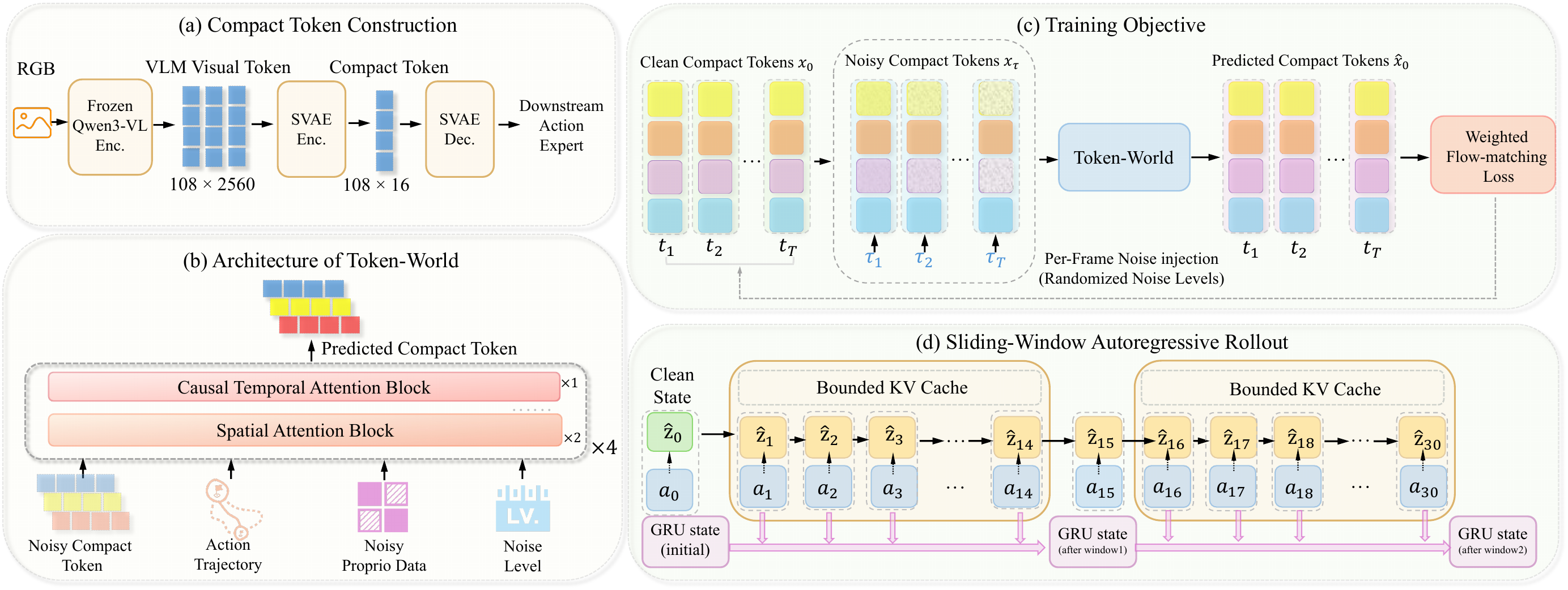}
\caption{
\textbf{Token-World overview.}
\textbf{(a) Compact Token Construction.} Qwen3-VL visual tokens are compressed into a compact latent space.
\textbf{(b) Dynamics Modeling.} A spatiotemporal Transformer models action-conditioned dynamics in compact token space.
\textbf{(c) Training Objective.} We train the model using an $x_0$-parameterized flow objective with weighted shortcut forcing.
\textbf{(d) Autoregressive Rollout.} Future compact tokens are generated with a sliding temporal window and recurrent state.
}
\label{fig:architecture}
\end{figure*}

\section{Token-World}
\label{sec:method}

\subsection{Overview}

Token-World separates the representation consumed by the policy from the state space modeled by the world model. Given policy-facing VLM tokens
$z_t\in\mathbb{R}^{N\times D_0}$, we first map each token into a compact representation
\begin{equation}
c_t=C_\phi(z_t)\in\mathbb{R}^{N\times d},
\qquad d\ll D_0,
\end{equation}
while preserving the number and spatial arrangement of tokens. Token-World then learns action-conditioned dynamics in this compact state space. Predicted compact states can be mapped back through a frozen decoder,
\begin{equation}
\hat z_t=G_\phi(\hat c_t),
\end{equation}
when policy-facing features are required for downstream control or evaluation.

The dynamics model is implemented as a spatio-temporal Transformer with factorized spatial and causal temporal attention \cite{vaswani2017attention,ho2019axial}, together with recurrent temporal memory for context beyond the finite attention window. We train the transition model using an $x_0$-parameterized flow objective with shortcut forcing \cite{hafner2025training,chen2024diffusion,frans2024one}, and perform autoregressive rollout directly in the compact state space.

\subsection{Compact VLM World State}

We use a separately trained semantic VAE (S-VAE) to construct the compact state space. The frozen VLM encoder produces
$z_t\in\mathbb{R}^{N\times D_0}$, and the compression encoder parameterizes a diagonal Gaussian posterior
\begin{equation}
q_\phi(c_t\mid z_t)
=
\mathcal{N}\left(
\mu_\phi(z_t),
\operatorname{diag}\sigma_\phi^2(z_t)
\right).
\end{equation}
During codec training, compact states are sampled using
\begin{equation}
c_t
=
\mu_\phi(z_t)
+
\sigma_\phi(z_t)\odot\epsilon,
\qquad
\epsilon\sim\mathcal{N}(0,I),
\end{equation}
while the posterior mean is used when exporting representations for world-model training.

The decoder reconstructs the original policy-facing representation as
\begin{equation}
\tilde z_t=G_\phi(c_t).
\end{equation}
Importantly, compression is applied along the feature dimension only: no spatial pooling or token fusion is performed, so the original spatial token layout is retained. In our main configuration, the VLM representation contains $N=108$ spatial tokens with $D_0=2560$ channels, which are compressed to $d=16$ channels per token.

The codec is optimized independently of the dynamics model using
\begin{equation}
\begin{aligned}
\mathcal{L}_{\mathrm{codec}}
={}&
\mathcal{L}_{\mathrm{MSE}}(\tilde z_t,z_t)
+
\mathcal{L}_{\mathrm{cos}}(\tilde z_t,z_t)
\\
&+
\lambda_{\mathrm{KL}}
D_{\mathrm{KL}}
\left(
q_\phi(c_t\mid z_t)
\,\|\,\mathcal{N}(0,I)
\right).
\end{aligned}
\end{equation}
where $\lambda_{\mathrm{KL}}$ controls the latent regularization. After training, both $C_\phi$ and $G_\phi$ are frozen when learning world dynamics.

\subsection{Action-Conditioned Dynamics Model}

We define the compact world state as
\begin{equation}
x_t=(c_t,p_t),
\end{equation}
where $p_t$ denotes the robot proprioceptive state. Token-World models the action-conditioned transition
\begin{equation}
p_\theta
\left(
x_{t+1}
\mid
x_{\leq t},a_t
\right).
\end{equation}

Compact visual tokens, proprioception, and actions are projected into a shared model space and processed by a spatio-temporal Transformer. Spatial attention captures interactions among tokens within each timestep, while causal temporal attention models their evolution over time. We factorize the two attention axes to avoid dense attention over the full space-time sequence \cite{ho2019axial}.

Because temporal attention operates over a finite context window, we additionally maintain a fixed-size GRU state \cite{cho2014learning} across temporal blocks. This recurrent state summarizes earlier interaction context that has fallen outside the active attention window and is propagated across successive rollout windows. We treat the Transformer and recurrent state as components of the transition model rather than as separate representation spaces.

\begin{figure*}[t]
    \centering
    \begin{subfigure}[t]{0.56\textwidth}
        \centering
        \includegraphics[width=\linewidth]{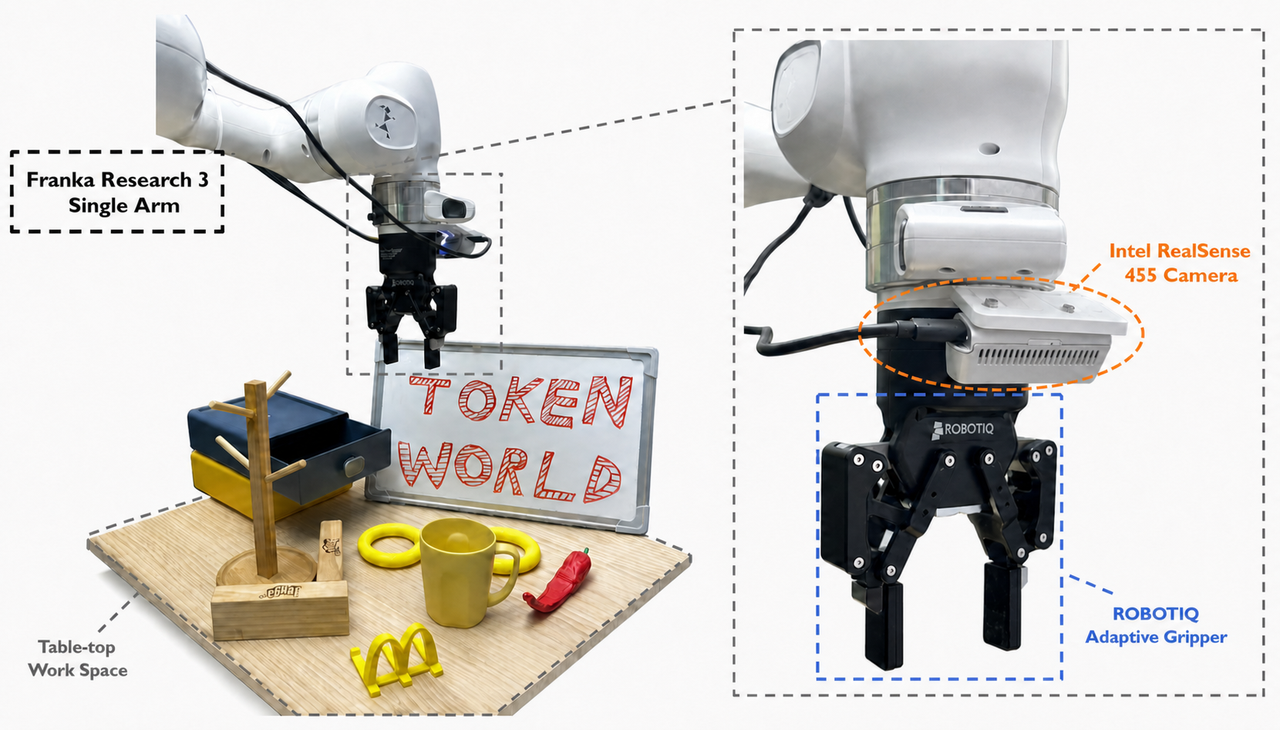}
        \caption{Table-top manipulation platform and wrist-mounted sensing setup.}
    \end{subfigure}
    \hfill
    \begin{subfigure}[t]{0.41\textwidth}
        \centering
        \includegraphics[width=\linewidth]{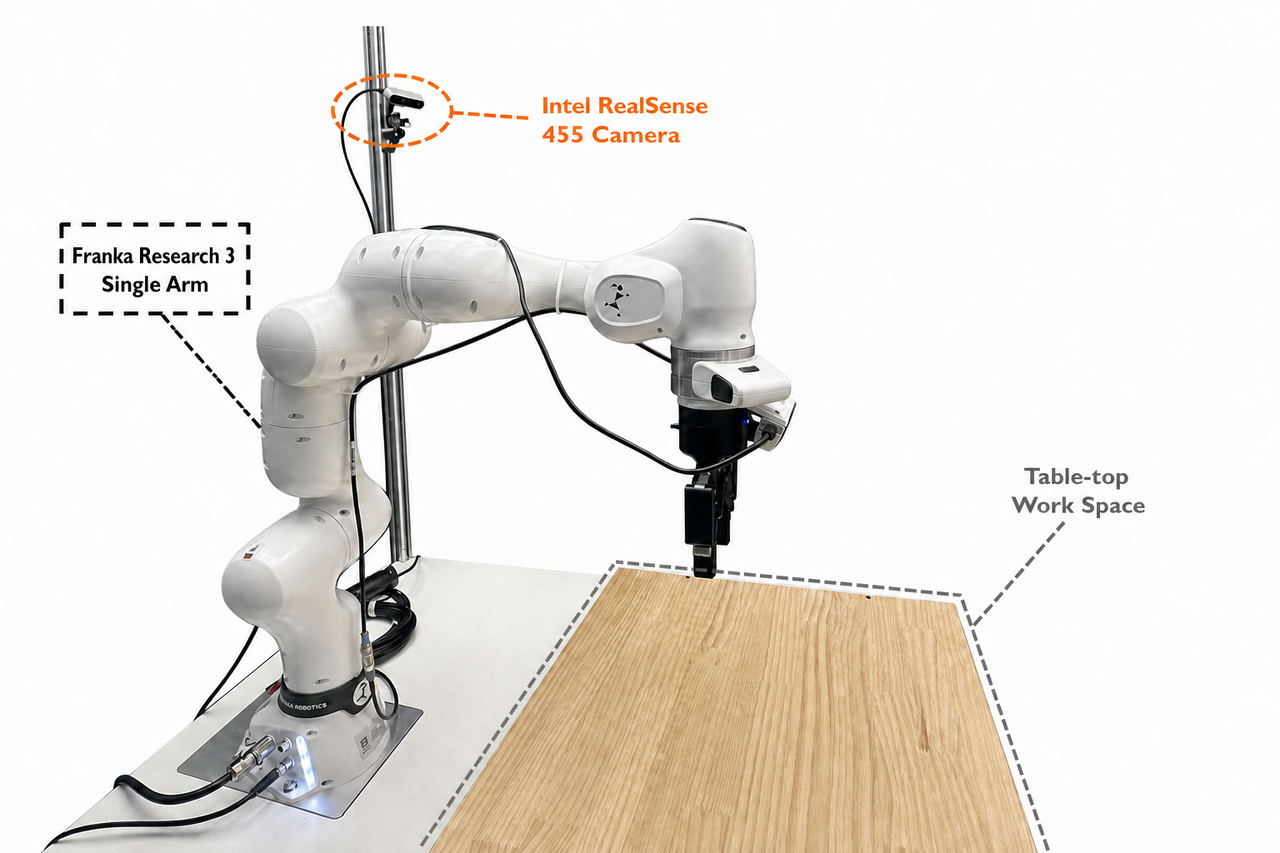}
        \caption{External view of the real-world data-collection setup.}
    \end{subfigure}
    \caption{
    \textbf{Real-world evaluation setup.}
    We collect manipulation trajectories using a Franka Research 3 arm equipped with a Robotiq adaptive gripper and two Intel RealSense 435 cameras.
    The real-world benchmark contains six manipulation tasks (hang on M, hang on cup, stack jenga, stack ring, put jenga in drawer, put chili in drawer) with 100 trajectories per task.
    }
    \label{fig:real_setup}
\end{figure*}

\subsection{Training and Autoregressive Rollout}

\paragraph{Flow-matching objective.}
Following \cite{hafner2025training,li2025back}, we train the dynamics model using an $x_0$-parameterized flow objective. Let $x$ denote a clean future compact world state and let
$\epsilon\sim\mathcal{N}(0,I)$. For flow time $\tau\in[0,1)$, we construct
\begin{equation}
x_\tau=(1-\tau)\epsilon+\tau x.
\end{equation}
The transition model predicts the clean endpoint conditioned on the available interaction history,
\begin{equation}
\hat x
=
f_\theta(x_\tau,\tau,\mathcal{C}),
\end{equation}
where $\mathcal{C}$ contains previous compact states, proprioception, actions, and recurrent context.

The endpoint prediction induces the flow
\begin{equation}
v_\theta
=
\frac{\hat x-x_\tau}{1-\tau},
\end{equation}
which is trained against the target flow using
\begin{equation}
\mathcal{L}_{\mathrm{flow}}
=
\mathbb{E}
\left[
w(\tau)
\left\|
v_\theta-v^\ast
\right\|_2^2
\right],
\qquad
w(\tau)=0.9\tau+0.1.
\end{equation}
We further adopt shortcut forcing \cite{hafner2025training,chen2024diffusion,frans2024one}, which trains the model on intermediate states generated from its own predictions and supports efficient few-step generation. The VLM encoder and semantic VAE remain frozen throughout world-model training.

\paragraph{Autoregressive rollout.}
At inference time, the observed compact state initializes the model context.
Future states are generated sequentially by denoising an initialized noisy state
conditioned on the current history and action. Each denoised prediction is
appended to the rollout history and used to condition subsequent transitions,
\begin{equation}
\hat x_{t+k+1}
\sim
p_\theta
\left(
\cdot
\mid
\hat x_{\leq t+k},
a_{t+k}
\right).
\end{equation}
Temporal attention is restricted to a bounded context window, while the recurrent
state is propagated across windows to retain longer-term interaction context.

The rollout remains entirely in the compact state space. When policy-facing
features are required, the predicted compact visual state is mapped back through
the frozen semantic decoder,
\begin{equation}
\hat z_{t+k}=G_\phi(\hat c_{t+k}),
\end{equation}
and the reconstructed VLM representation is consumed directly by the downstream
policy. No RGB reconstruction is involved in the autoregressive dynamics or
policy-feedback loop.

\paragraph{RGB rendering for evaluation.}
For qualitative visualization and manual task-success assessment, a separately
trained and frozen convolutional decoder $D_{\mathrm{rgb}}$ reconstructs
$224\times224$ RGB frames from predicted VLM visual-tokens.
This decoder is used only for rendering and does not participate in world-model
rollout or policy inference.

\section{Experiments}
\label{sec:experiments}

We evaluate Token-World from three complementary perspectives: open-loop prediction fidelity, closed-loop policy evaluation, and representation design. We first compare Token-World with recent robotic world models and analyze how prediction quality changes over long rollout horizons. We then test whether these gains translate into more reliable and efficient closed-loop policy simulation by comparing simulated and reference success rates. Finally, controlled ablations study how the design and dimensionality of the compact S-VAE state affect dynamics prediction.

\begin{table*}[t]
\centering
\caption{
\textbf{Open-loop world-model fidelity on simulated and real-world manipulation.}
Feature metrics compare predicted and ground-truth frozen VLM representations.
Policy-action metrics compare the outputs of the same frozen policy conditioned on predicted and ground-truth representations.
For RGB-generative baselines, generated frames are re-encoded by the frozen VLM before evaluation.
Higher cosine similarity and lower NMSE are better.
}
\label{tab:main_benchmark}

\setlength{\tabcolsep}{5.5pt}
\renewcommand{\arraystretch}{1.18}
\fontsize{8.6pt}{10.5pt}\selectfont

\begin{tabular*}{\textwidth}{
@{\extracolsep{\fill}}
l
c
cccc
cccc
@{}
}
\toprule

\multirow{3}{*}{\textbf{Method}}
&
\multirow{3}{*}{\textbf{Venue}}
&
\multicolumn{4}{c}{\textbf{RoboTwin (50 Tasks)}}
&
\multicolumn{4}{c}{\textbf{Real Franka (6 Tasks)}}
\\

\cmidrule(lr){3-6}
\cmidrule(lr){7-10}

&
&
\multicolumn{2}{c}{\textbf{VLM Feature Fidelity}}
&
\multicolumn{2}{c}{\textbf{Policy-Action Consistency}}
&
\multicolumn{2}{c}{\textbf{VLM Feature Fidelity}}
&
\multicolumn{2}{c}{\textbf{Policy-Action Consistency}}
\\

&
&
\textbf{Cosine}$\uparrow$
&
\textbf{NMSE}$\downarrow$
&
\textbf{Cosine}$\uparrow$
&
\textbf{NMSE}$\downarrow$
&
\textbf{Cosine}$\uparrow$
&
\textbf{NMSE}$\downarrow$
&
\textbf{Cosine}$\uparrow$
&
\textbf{NMSE}$\downarrow$
\\
\midrule

IRASim
& ICCV'25
& 0.7372
& 0.5685
& 0.9260
& 0.1533
& 0.7061 & 0.6095 & 0.9174 & 0.1702
\\

Ctrl-World
& ICLR'26
& 0.7097
& 0.6437
& 0.9265
& 0.1463
& 0.7213 & 0.5719 & 0.9288 & 0.1410
\\

WorldGym
& ICLR'26
& 0.7146 & 0.6157 & 0.9251 & 0.1602
& 0.6697 & 0.6930 & 0.9089 & 0.1981
\\

\midrule

\textbf{Token-World}
& \textbf{Ours}
& \textbf{0.7714}
& \textbf{0.4892}
& \textbf{0.9439}
& \textbf{0.1145}
& \textbf{0.7506} & \textbf{0.5129} & \textbf{0.9402} & \textbf{0.1208}
\\

\bottomrule
\end{tabular*}


\end{table*}

\subsection{Experimental Setup}
\label{subsec:exp_setup}

\paragraph{Datasets.}
We evaluate Token-World on both simulated and real-world manipulation data.
For simulation, we use 50 manipulation tasks from RoboTwin~\cite{chen2025robotwin}, with 100 trajectories per task, including 50 successful and 50 failed demonstrations, for a total of 5,000 trajectories.
For real-world evaluation, we collect a Franka manipulation dataset containing six tasks and 600 successful trajectories.
Both datasets use a 9:1 train--validation split, and all compared methods share the same training and evaluation data.


\begin{figure*}[t]
    \centering
    \includegraphics[width=\textwidth]
    {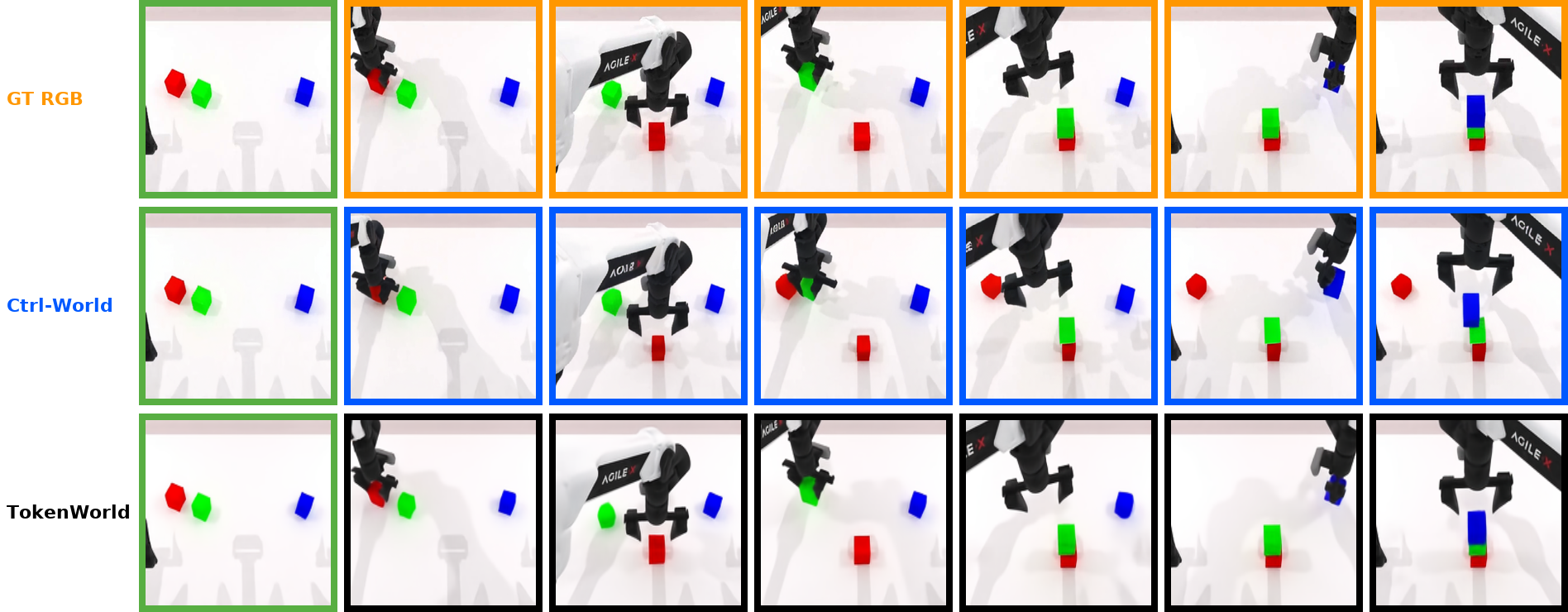}
    \caption{
    \textbf{Qualitative comparison of long-horizon open-loop rollout.}
    We visualize seven states along the same action-replay trajectory.
    The top row shows the reference RGB observations, followed by predictions from Ctrl-World and Token-World.
    Token-World more closely preserves the task-relevant object configuration and interaction progression over long horizons.
    Token-World predictions are rendered to RGB using the auxiliary decoder only for visualization.
    }
    \label{fig:openloop_rollout_vis}
\end{figure*}

\begin{figure}[t]
    \centering
    \includegraphics[width=0.85\linewidth]
    {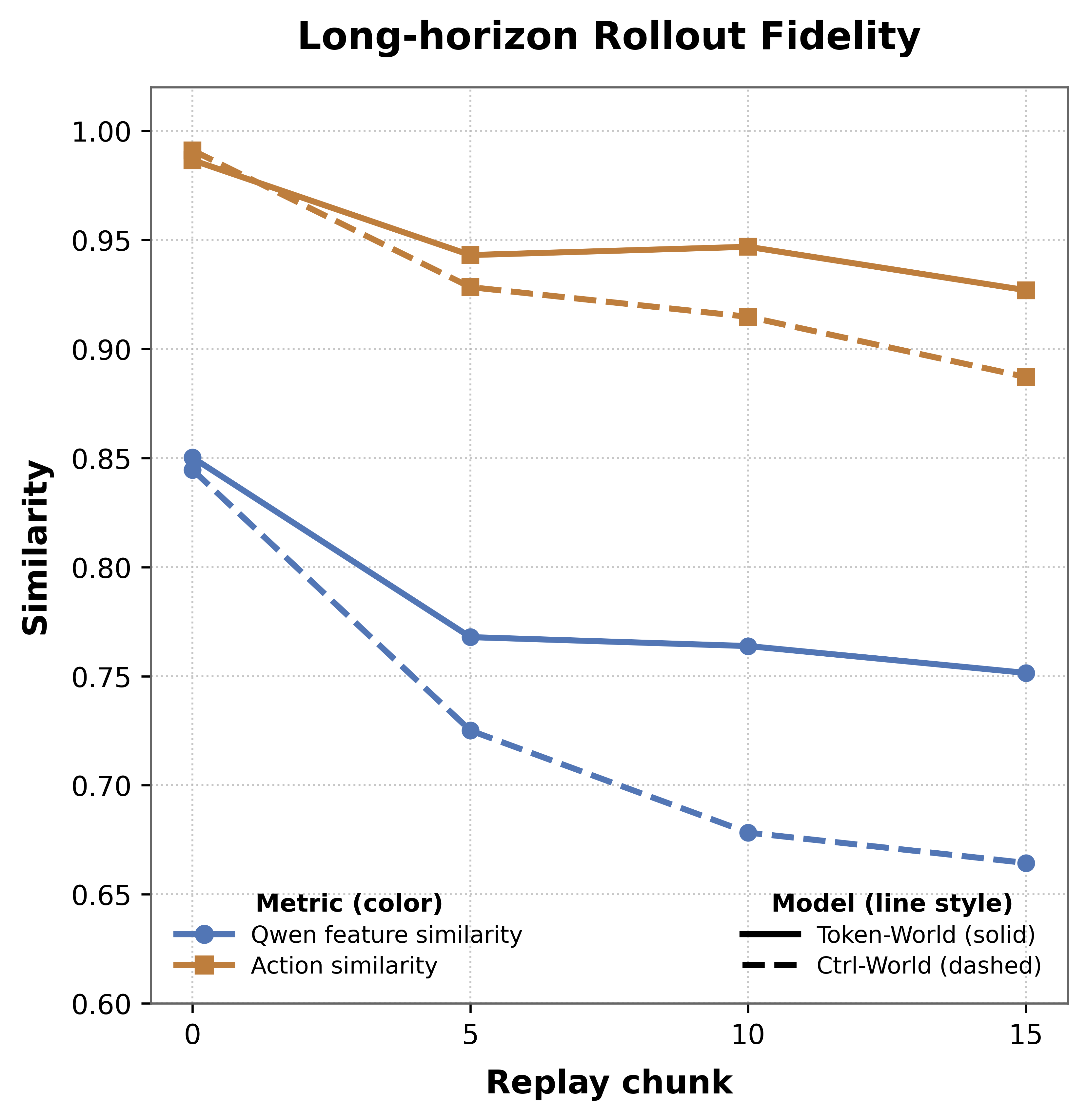}
    \caption{
    \textbf{Long-horizon open-loop fidelity under chunk-based autoregressive rollout.}
    Results are reported at replay chunks 0, 5, 10, and 15, with 16 steps a chunk.
    Each point denotes the chunk-wise average VLM-feature and action similarity.
    Token-World degrades more slowly than Ctrl-World over long-horizon rollout.
    }
    \label{fig:horizon_curve}
\end{figure}

\paragraph{Baselines.}
We compare Token-World with three recent robotic world models:
IRASim~\cite{zhu2025irasim} (ICCV 2025),
Ctrl-World~\cite{guo2025ctrl} (ICLR 2026),
and WorldGym~\cite{quevedo2025worldgym} (ICLR 2026).
These methods represent competitive learned simulators based on generative visual dynamics.


\paragraph{Open-loop evaluation.}
Under open-loop action replay, each world model predicts future states from the
recorded action sequence. We evaluate VLM-feature fidelity using NMSE and cosine
similarity. To measure policy-action consistency, we feed the predicted and
ground-truth representations separately to the same frozen policy and compare
the resulting actions. Metrics are averaged over
trajectories within each task and then macro-averaged across tasks.


\paragraph{Closed-loop evaluation and efficiency.}
For closed-loop policy evaluation, we use three StarVLA variants
(StarVLA-OFT, StarVLA-PIv3, and StarVLA-GR00T)
\cite{community2026starvla}, which share the same Qwen3-VL
\cite{bai2025qwen3} backbone but use different action heads.
Each rollout is initialized from the same recorded state as its reference
environment counterpart, after which predicted states are fed back to the policy
to generate subsequent actions. We compare simulated and reference success rates
using Pearson correlation, with success manually assessed from decoded RGB
rollouts.

We measure batch-1 latency on PPU-ZW810E accelerators using the same inference
settings as the fidelity evaluations: RGB baselines use 50 denoising steps,
whereas Token-World uses 16 steps with shortcut forcing. Latency is reported per
predicted observation to account for different numbers of future observations
generated per diffusion call. Timing includes RGB decoding and VLM re-encoding
for RGB baselines, and compact-state prediction plus direct decoding to
policy-facing VLM features for Token-World.


\begin{figure*}[t]
    \centering

    \includegraphics[
        width=\textwidth
    ]{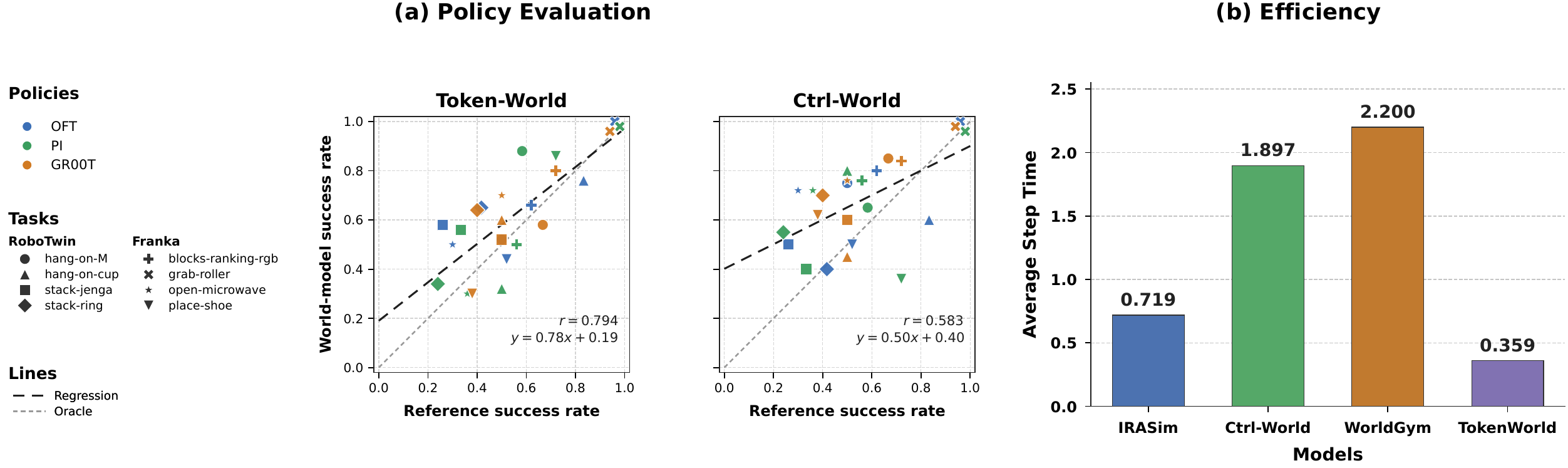}

    \vspace{-0.6em}

    \caption{
        \textbf{Policy evaluation fidelity and simulation efficiency.}
        \textbf{(a)} Correlation between policy success rates measured
        in learned world models and the corresponding reference environments.
        Each point represents one policy--task pair.
        The gray dotted line indicates the oracle relation $y=x$, and the
        black dashed line denotes linear regression.
        \textbf{(b)} Average per-step simulation time of different
        world-model simulators. Lower is better.
    }
    \label{fig:policy_evaluation_efficiency}
\end{figure*}

\subsection{Open-Loop Prediction Fidelity}
\label{subsec:open_loop}

We first evaluate Token-World under open-loop action replay, considering both overall prediction fidelity and long-horizon degradation.

\paragraph{Overall open-loop fidelity.}
Table~\ref{tab:main_benchmark} compares Token-World with recent robotic world models.
On RoboTwin, Token-World achieves a feature cosine similarity of $0.7714$ and NMSE of $0.4892$, compared with $0.7372/0.5685$ for IRASim and $0.7097/0.6437$ for Ctrl-World.
The gains persist after policy inference, with an action cosine similarity of $0.9439$ and NMSE of $0.1145$, compared with $0.9260/0.1533$ and $0.9265/0.1463$, respectively.


\paragraph{Long-horizon rollout fidelity.}
We compare Token-World and Ctrl-World at replay chunks $0$, $5$, $10$, and $15$, with 16 actions per chunk. Figure~\ref{fig:horizon_curve} reports the corresponding chunk-wise Qwen-feature and action similarity.

\paragraph{Qualitative long-horizon rollout.}
Figure~\ref{fig:openloop_rollout_vis} provides a qualitative comparison over a long open-loop rollout. Token-World better preserves task-relevant object configurations and robot--object interactions across the trajectory, while Ctrl-World gradually deviates from the reference evolution at later steps. For visualization, Token-World predictions are decoded from the compact token space back to RGB; RGB reconstruction is not used during the rollout itself.

\subsection{Closed-Loop Policy Evaluation}
\label{subsec:closed_loop}

Open-loop evaluation measures prediction under fixed action sequences, while closed-loop simulation tests whether these predictions remain useful once they influence subsequent policy actions. We therefore evaluate Token-World as a policy simulator by measuring whether it preserves task-level outcomes under closed-loop rollout.

\paragraph{Success-rate correlation.}
Figure~\ref{fig:policy_evaluation_efficiency} compares simulated and reference success rates across policy--task pairs. Token-World increases the Pearson correlation from $0.583$ to $0.794$ over Ctrl-World, with a regression closer to the oracle $y=x$. This indicates that its open-loop fidelity gains translate into more reliable closed-loop policy evaluation.

\paragraph{Simulation efficiency.}
As shown in Fig.~\ref{fig:policy_evaluation_efficiency}, Token-World achieves
the lowest per-step simulation time at $0.359$\,s, yielding $2.0\times$,
$5.3\times$, and $6.1\times$ speedups over IRASim, Ctrl-World, and WorldGym, respectively.

\subsection{Representation Design Ablations}
\label{subsec:representation_ablation}

We further investigate the source of these gains through dimensionality studies, dynamics validation, and a controlled S-VAE versus VAE comparison.

\paragraph{Reconstruction--modelability trade-off.}
We first vary the channel dimension of the S-VAE while preserving the same spatial token structure.
For each representation, we measure both reconstruction fidelity and diffusion-modelability properties.

\begin{table}[t]
\centering
\caption{
\textbf{Reconstruction and modelability of S-VAE representations across compact-state dimensions.}
Increasing the latent dimension improves reconstruction fidelity, whereas the modelability metrics exhibit the opposite trend.
}
\label{tab:svae_representation}

\setlength{\tabcolsep}{4.2pt}
\renewcommand{\arraystretch}{1.10}

\begin{tabular}{@{}ccccc@{}}
\toprule
\textbf{$d$}
& \textbf{NMSE}$\downarrow$
& \textbf{Cos.}$\uparrow$
& \textbf{SEC}$\downarrow$
& \textbf{LDS}$\uparrow$
\\
\midrule

8
& 0.1026
& 0.9389
& 0.3511
& 0.3100
\\

16
& 0.0791
& 0.9528
& 0.3854
& 0.2534
\\

24
& 0.0694
& 0.9587
& 0.3960
& 0.2383
\\

32
& 0.0633
& 0.9623
& 0.4058
& 0.2274
\\

48
& 0.0562
& 0.9668
& 0.4269
& 0.1969
\\

96
& 0.0445
& 0.9739
& 0.4394
& 0.1595
\\

128
& 0.0409
& 0.9762
& 0.4740
& 0.1182
\\

\bottomrule
\end{tabular}


\end{table}

Following prior analyses of latent diffusability, we use Spectral Energy
Concentration (SEC)~\cite{zhong2026diffusing} and Local-vs-Distant Similarity
(LDS)~\cite{singh2025matters} to characterize spectral smoothness and spatial
structure, respectively. Lower SEC indicates smoother latent features, while
higher LDS indicates stronger local spatial structure. As shown in
Table~\ref{tab:svae_representation}, increasing the compact dimension from
$8$ to $128$ improves reconstruction (NMSE: $0.1026\!\rightarrow\!0.0409$;
Cos.: $0.9389\!\rightarrow\!0.9762$), but increases SEC
($0.3511\!\rightarrow\!0.4740$) and decreases LDS
($0.3100\!\rightarrow\!0.1182$). This suggests that higher reconstruction
fidelity does not necessarily correspond to more diffusion-friendly latent
properties.

\paragraph{Effect of compact-state dimension on dynamics prediction.}
Representation-level diagnostics alone do not establish whether the observed trend matters for an actual world model.
We therefore train the same dynamics backbone using S-VAE representations with different compact dimensions.
The architecture, training data, optimization protocol, and evaluation procedure are held fixed; only the compact-state dimensionality is varied.

\begin{table}[t]
\centering
\caption{
\textbf{Effect of S-VAE dimensionality on downstream dynamics prediction.}
The dynamics backbone, training data, and optimization protocol are fixed;
only the compact-state dimension is varied.
}
\label{tab:dim_ablation}

\setlength{\tabcolsep}{4.0pt}
\renewcommand{\arraystretch}{1.08}

\begin{tabular}{@{}lcccc@{}}
\toprule
\textbf{Representation}
& \textbf{Feat. NMSE}$\downarrow$
& \textbf{Feat. Cos.}$\uparrow$
& \textbf{Act. NMSE}$\downarrow$
& \textbf{Act. Cos.}$\uparrow$
\\
\midrule

S-VAE-8
& 0.1549
& 0.8254
& 0.0259
& 0.9603
\\

S-VAE-16
& \textbf{0.1517}
& \textbf{0.8306}
& 0.0247
& \textbf{0.9641}
\\

S-VAE-32
& 0.1634
& 0.8182
& 0.0275
& 0.9596
\\

S-VAE-48
& 0.1698
& 0.8109
& \textbf{0.0236}
& 0.9638
\\

\bottomrule
\end{tabular}

\end{table}

\begin{table}[t]
\centering
\caption{
\textbf{Controlled ablation of world-state representations.}
All variants use the same spatiotemporal dynamics backbone and training protocol;
only the state representation and its corresponding input/output projections are changed.
}
\label{tab:representation_ablation}

\setlength{\tabcolsep}{3.8pt}
\renewcommand{\arraystretch}{1.08}

\begin{tabular}{@{}lcccc@{}}
\toprule
\textbf{Representation}
& \textbf{Feat. NMSE}$\downarrow$
& \textbf{Feat. Cos.}$\uparrow$
& \textbf{Act. NMSE}$\downarrow$
& \textbf{Act. Cos.}$\uparrow$
\\
\midrule

Raw VLM
& 0.7851
& 0.6375
& 0.2180
& 0.8948
\\

SDXL-VAE
& 0.8164
& 0.621
& 0.2586
& 0.875
\\

S-VAE
& \textbf{0.4892}
& \textbf{0.7714}
& \textbf{0.1145}
& \textbf{0.9439}
\\

\bottomrule
\end{tabular}
\end{table}

Among the evaluated dimensions, $d=16$ provides the best overall trade-off for dynamics prediction. It achieves the lowest feature NMSE ($0.1517$), highest feature cosine similarity ($0.8306$), and highest action cosine similarity ($0.9641$). Reducing the dimension further to $8$ slightly degrades both feature and action consistency, while larger dimensions $32$ and $48$ yield worse future-feature prediction. These results indicate that dynamics performance is non-monotonic with representation capacity, supporting $d=16$ as the operating point for Token-World.

\paragraph{Effect of world-state representation.}
We finally examine whether compact S-VAE states are easier to model than either
the original high-dimensional VLM features or generic image-VAE latents.
We compare raw Qwen3-VL visual features~\cite{bai2025qwen3}, SDXL-VAE latents~\cite{podell2024sdxl},
and our compact S-VAE states under the same dynamics backbone and training protocol.

As shown in Table~\ref{tab:representation_ablation}, raw VLM features already
outperform SDXL-VAE latents, but remain substantially worse than S-VAE.
Compared with raw VLM features, S-VAE reduces feature/action NMSE from
$0.7851/0.2180$ to $0.4892/0.1145$, while improving cosine similarity from
$0.6375/0.8948$ to $0.7714/0.9439$.

\section{Conclusion}
\label{sec:conclusion}

We presented Token-World, an autoregressive world-model simulator that learns action-conditioned dynamics in a compact policy-facing VLM token space. By compressing high-dimensional VLM features into a dynamics-friendly state, Token-World avoids intermediate RGB generation while retaining compatibility with downstream VLA policies. Experiments on simulated and real-world manipulation show improved open-loop feature and policy-action fidelity, more stable long-horizon rollouts, stronger agreement with reference policy performance in closed-loop evaluation, and lower simulation latency than recent world-model baselines. Representation ablations further highlight the importance of compact-state design for dynamics modeling. Our current evaluation focuses on manipulation tasks and policies sharing a common VLM backbone, while we do not yet systematically characterize what makes a semantic representation suitable for world modeling. Extending Token-World across policy backbones, interaction data, and representation designs remains important future work.









\bibliographystyle{IEEEtran}
\bibliography{ref}

\end{document}